\pdfoutput=1

\documentclass[11pt]{article}

\usepackage[compact]{titlesec}
\titlespacing{\section}{0pt}{*1}{*0.5}
\titlespacing{\subsection}{0pt}{*0.5}{*0.3}

\usepackage{enumitem}
\setlist{noitemsep}

\usepackage{longtable} 

\usepackage[final]{coling}

\usepackage{times}
\usepackage{latexsym}

\usepackage{tabularx} 
\usepackage{multirow}
\usepackage{array} 

\usepackage[T1]{fontenc}
\usepackage[utf8]{inputenc}
\usepackage{microtype}
\usepackage{inconsolata}
\usepackage{graphicx}

\usepackage{amsmath}
\usepackage{amssymb}

\title{Assessing the Answerability of Queries in Retrieval-Augmented \\ Code Generation}

\author{
\begin{tabular}{@{}c c}
Geonmin Kim\thanks{Equal contribution.}\quad
Jaeyeon Kim\footnotemark[1]\quad
Hancheol Park\quad\\
Wooksu Shin\quad
Tae-Ho Kim\thanks{Corresponding author.}\vspace{6pt} 
\end{tabular}\\
Nota Inc.\\
\texttt{\{geonmin.kim, jaeyeon.kim, hancheol.park, wooksu.shin, thkim\}@nota.ai}
}

\begin{document}
\maketitle
\begin{abstract}
Thanks to unprecedented language understanding and generation capabilities of large language model (LLM), Retrieval-augmented Code Generation (RaCG) has recently been widely utilized among software developers. While this has increased productivity, there are still frequent instances of incorrect codes being provided. In particular, there are cases where plausible yet incorrect codes are generated for queries from users that cannot be answered with the given queries and API descriptions. This study proposes a task for evaluating answerability, which assesses whether valid answers can be generated based on users' queries and retrieved APIs in RaCG. Additionally, we build a benchmark dataset called Retrieval-augmented Code Generability Evaluation (RaCGEval) to evaluate the performance of models performing this task. Experimental results show that this task remains at a very challenging level, with baseline models exhibiting a low performance of 46.7\%. Furthermore, this study discusses methods that could significantly improve performance.

\end{abstract}

\begin{figure}[t!]
  \includegraphics[width=\columnwidth]{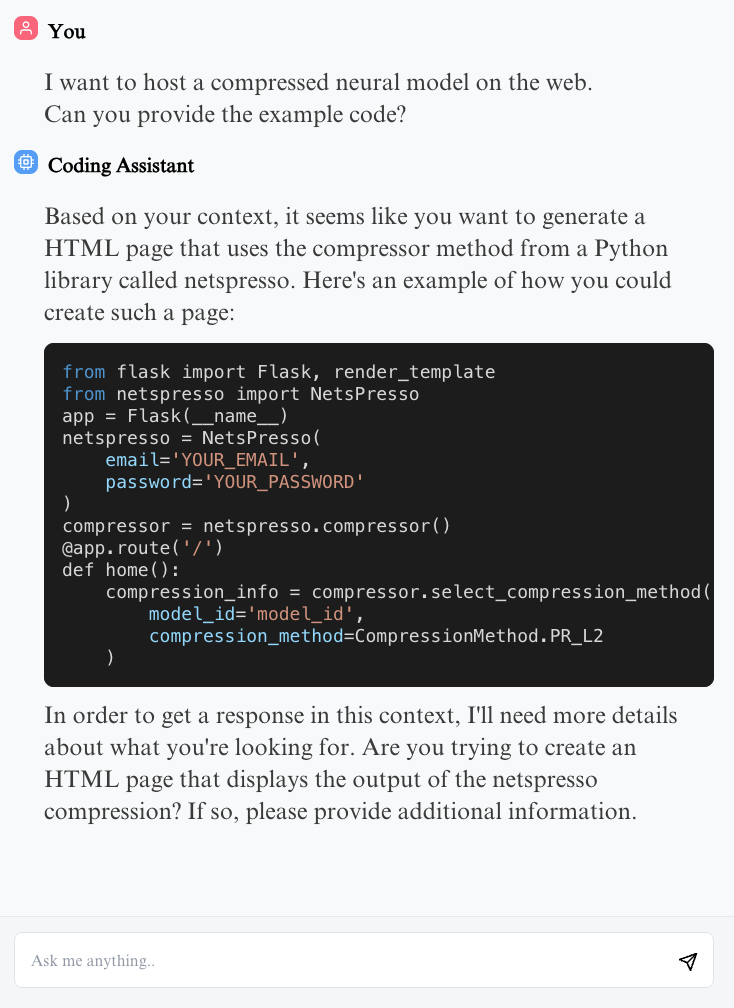}
  \caption{An example of an LLM generating plausible code even when the request is made outside the functionality provided by the library. When generating code based on the API documentation of NetsPresso~\cite{netspresso2024}, a deep learning model optimization library, an LLM generates the response to a query requesting web page creation using Netspresso API.}
  \label{fig:1}
\end{figure}
\section{Introduction}
Recently, the large language models (LLMs) have demonstrated unprecedented performance on a variety of natural language processing tasks. Based on their language understanding and generation capabilities, they are now widely used as coding assistants to support developers across various industries. This has significantly improved software development productivity. In particular, Retrieval-augmented Generation (RAG)~\cite{shuster2021RAGreducehalu} has enabled LLMs to generate code for various libraries that they were not trained on.
As depicted in Figure~\ref{fig:1}, when generating code through Retrieval-augmented Code Generation (RaCG), LLMs often produce plausible but incorrect code for a user's query, which may not be answerable. Developers using these code assistants in a RaCG environment may often ask questions that fall outside the capabilities provided by the library because they are not proficient in using it. Therefore, in such cases, LLMs must be able to determine whether they can provide an answer before generating a response code for the user's query.

\begin{figure*}[t!]
    \centering
    \includegraphics[width=1.0\linewidth]{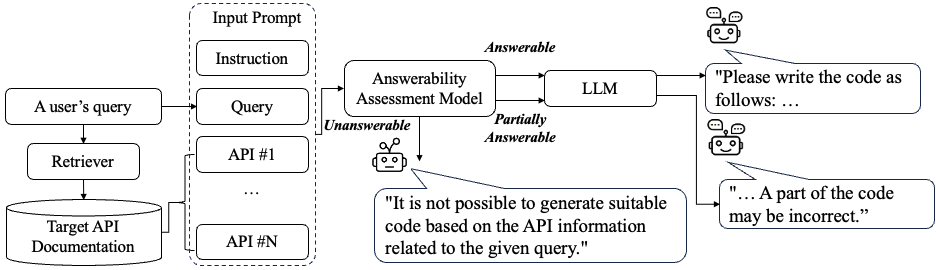}
    \caption{The process of evaluating the answerability of a user's query in RaCG and generating code accordingly or rejecting code generation}
    \label{fig:2}
\end{figure*}

This can be seen as a well-known hallucination problem. Although RAG is known to mitigate the hallucination by making the language model's response grounded on external knowledgebase, the hallucination still exists~\cite{esetal2024ragas,li-etal-2024-llatrieval}. To reduce such hallucinations, various methods for detecting hallucinations have been studied. SelfCheckGPT~\cite{manakul2023selfcheckgpt} uses the consistency of multiple sampled responses to determine whether generated responses include hallucination. This method requires the sampling of a large number of responses to determine hallucination. RAGAS~\cite{esetal2024ragas} and LLatriEval ~\cite{li-etal-2024-llatrieval} examine whether the responses generated in a RAG scenario are well-supported by the documents. However, these methods all require the generation of responses by the LLM one or multiple times. 
As an alternative to approach hallucination, one can think of assessing the answerability of the input prompt before LLM generates the response. There has not been much research on this task in the RaCG scenario. If it is possible to evaluate answerability without generating multiple responses, it could reduce computational costs for LLM service providers. Additionally, users would be able to reduce their waiting time for responses and modify their queries to receive the desired answers more quickly. This study discusses the task of evaluating such answerability.

Moreover, various benchmarks have been proposed to evaluate hallucinations in LLMs including general \cite{li2023halueval} and code domain~\cite{tian2024codehalu}. Most work addresses the issue of determining whether the generated responses are hallucination. However, to the best of our knowledge, there is no dataset available for assessing whether the given queries themselves are answerable. In this study, we build and release a benchmark dataset, \textbf{R}etrieval-\textbf{a}ugmented \textbf{C}ode \textbf{G}enerability \textbf{Eval}uation (RaCGEval), annotated for whether the queries given in RaCG are answerable.

Our contributions can be summarized as follows:
\begin{itemize}
  \item We propose methods to construct a benchmark for assessing the answerability of users' queries in RaCG. 
  \item We publicly release the RaCGEval benchmark. 
  \item We built models to evaluate the answerability of given users' queries and conducted an in-depth investigation on this task.  
\end{itemize}
RaCGEval benchmark dataset and pre-trained answerability assessment models are publicly available. \footnote{ \url{https://github.com/Nota-NetsPresso/RaCGEval}}
\section{Constructing a Benchmark Dataset}
In this section, we provide a detailed explanation of our task, which involves assessing whether a given user's query is answerable in RaCG (\textsection 2.1). We also describe how we constructed the benchmark dataset, RaCGEval, to evaluate the performance of models in solving this task (\textsection 2.2, \textsection 2.3, and \textsection 2.4), and discuss the reliability of this dataset (\textsection 2.4).

\begin{figure*}[t!]
    \centering
    \includegraphics[width=1.0\linewidth]{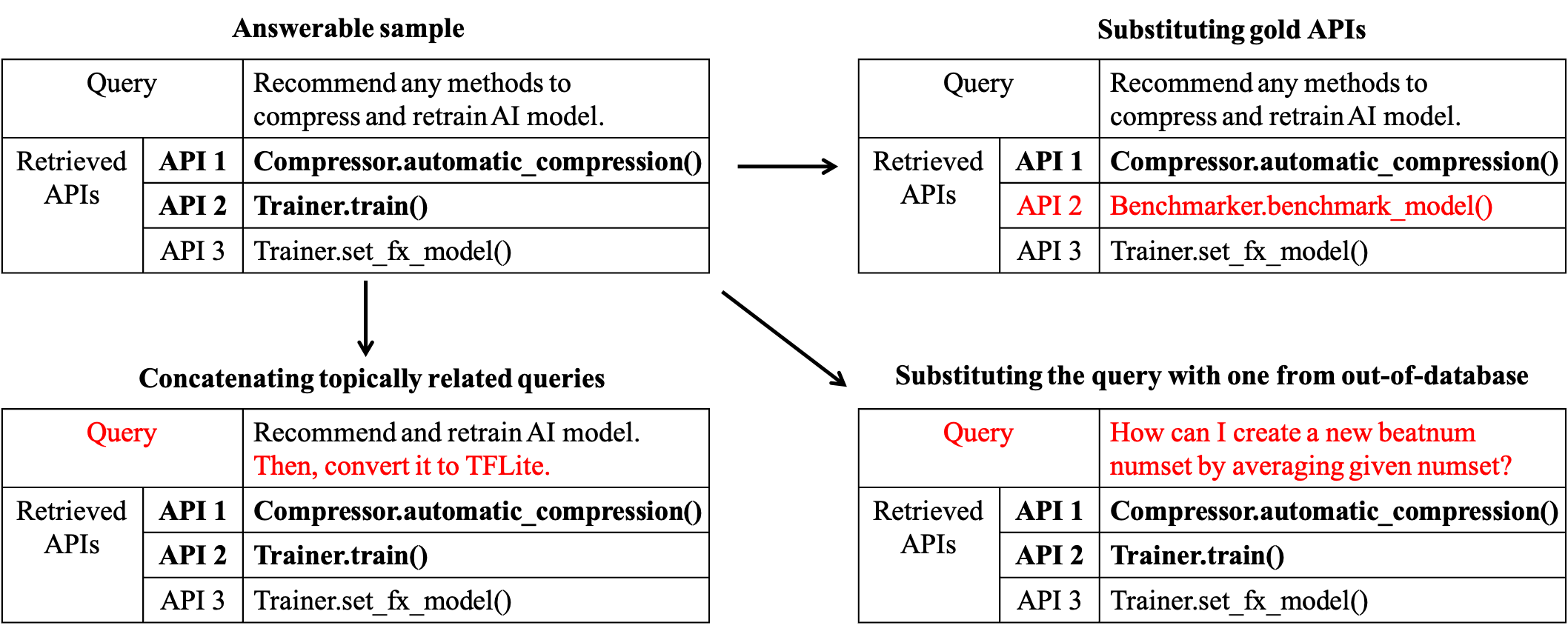}
    \caption{Examples of NegGen method to build unanswerable/partially answerable samples. The API highlighted in bold is the gold API for the given query in the answerable set.}
    \label{fig:3}
\end{figure*}

\subsection{Task description}
\label{sec:2_1}
Assessing the answerability of users' queries in the RaCG involves determining whether correct codes can be derived solely from the information provided in the input prompts. As depicted in Figure~\ref{fig:2}, each input prompt is created by combining a user's query with relevant APIs extracted from the target API documentation associated with that query. Each query typically consists of a single request, but it can sometimes contain multiple requests that share similar topics. Accordingly, if all requests can be answered through the extracted API descriptions, the query can be considered \textbf{answerable}. If only some requests can be resolved, it can be deemed \textbf{partially answerable}. If none of the requests can be resolved, it can be regarded as \textbf{unanswerable}. In conclusion, our task can be viewed as a three-way classification problem that takes an input prompt (i.e., instruction for answerability assessment with pairs of a user query and API descriptions) as input.

\subsection{API documentation}
\label{sec:2_2}
In our task, the language model must be able to determine whether an answer can be provided based solely on the information in the given input prompt. Therefore, to ensure a fair assessment that excludes prior knowledge of each LLM, the benchmark dataset includes APIs that most language models are unlikely to have encountered during training. Therefore, we use the following datasets that contain private or modified API information:

\begin{itemize}
  \item NetsPressoEval~\cite{netspresso2024}: The API documentation of NetsPresso, a deep learning model optimization platform, includes functions for compressing deep learning models or compiling them for specific hardware
  \item TorchDataEval~\cite{zan-etal-2022-language}: A collection of API descriptions for TorchData, which is a beta library for a modular data loading framework and efficient data pipelines
  \item BeatNumEval~\cite{zan-etal-2022-language}: The API documentation for Python NumPy, with its keywords and structure rephrased
  \item MonkeyEval~\cite{zan-etal-2022-language}: The API documentation for Python Pandas, with its keywords and structure rephrased
\end{itemize}
The other three datasets, excluding NetsPressoEval, are datasets studied in previous research~\cite{zan-etal-2022-language, zan2023privatelibrary}. Each sample in these datasets consists of a user's query and the retrieved APIs with a description. We manually add gold APIs, which can address user queries, if not presented in retrieved APIs. Each sample can be viewed as an answerable input prompt. Unlike the other datasets, NetsPressoEval is a newly created dataset for this study, and the answerable input prompts were generated by consulting those who develop the APIs. An example of the dataset for NetsPressoEval can be found in Table~\ref{tab:3} in Appendix~\ref{appendix:a}. 

\subsection{Generating partially answerable and unanswerable samples}
\label{sec:2_3}

So far, we have only been able to obtain answerable samples. Here, we propose a method to generate partially answerable and unanswerable samples using the answerable samples.

Currently, each answerable sample consists of a query ($q^*$)
and retrieved APIs ($\hat{R}=\{\hat{r}_1,\hat{r}_2,...\}$). The retrieved APIs also include gold APIs ($R^*=\{r_1^*, r_2^*, ...\}$). In this study, all samples are limited to having three retrieved APIs for a single query. This is to account for the input length limitations of many language models (i.e., 8192 tokens for llama-3-7B) and prompt length increases with the number of APIs included. The APIs associated with each query are determined based on their similarity scores to the query, in descending order. We denote the final selected APIs as $R=\{r_1, r_2, r_3\}$. We generated partially answerable and unanswerable samples using the following methods:
\begin{itemize}
  \item Substituting the gold APIs: To generate partially answerable and unanswerable samples, we substitute $r_i^* \in R^*$ with $\hat{r_j}$ such that $S(q^*, \hat{r_j})\geq\tau_{r}$, $\hat{r_j} \notin R^*$ and $\hat{r_j} \notin R$. $S( , )$ is the similarity score between two descriptions, which is measured in our study using the encoder of CodeT5+~\cite{wang-etal-2023-codet5}. $\tau_{r}$ and $\tau_{i}$ are the relevant and irrelevant thresholds. In this study, we use $\tau_{r}=0.8$, and $\tau_{i}=0.1$. In other words, it involves replacing the gold APIs with APIs that are semantically similar to the query but do not assist in generating the code corresponding to that query. If all gold APIs are replaced, the sample becomes an unanswerable sample. However, if only some are replaced, it becomes a partially answerable sample because certain request requirements in the query cannot be fulfilled.
  \item Concatenating topically related queries: To obtain partially answerable samples, we concatenate $q^*$ with $q$ such that $S(q,q^*)\geq\tau_{r}$ and $S(q,r^*)<\tau_{i}$. Here, $q$ refers to one of the queries included in the dataset related to the same API documentation. The newly added queries are unrelated to the gold APIs and therefore cannot be resolved. As a result, these samples become partially unanswerable samples.   
  \item Substituting the query with one from out-of-database: To generate unanswerable samples, we substitute $q^*$ with $q$ such that $S(q,r^*) < \tau_{i}$. The $q$ comes from other datasets.
\end{itemize}
Our methods are well illustrated in Figure~\ref{fig:3}. 

\subsection{Annotation}
Since noise may be included not only in answerable samples but also in unanswerable samples, we requested four programming experts to individually annotate labels related to answerability. An example of noise is when the gold APIs are replaced, resulting in a partially answerable sample, but the substituted APIs also help in generating the correct code. After the four annotators completed their annotations, a discussion session was held for the samples that did not reach a unanimous agreement in order to determine the final labels.

To verify whether the annotated dataset from four experts aligns with the opinions of other experts, annotations were requested from an additional two experts. These two experts had no access to the previously annotated information and performed their annotations independently. In total, we obtained three sets of labels, including the previously annotated data. We measured the inter-annotator agreement using Fleiss' Kappa~\cite{fleiss1971agreement}. As a result, we get a kappa score of 0.7408 which is interpreted as substantial agreement~\cite{landis1977kappascore}. Therefore, the annotation can be considered fairly reliable. The statistics of the RaCGEval benchmark dataset is described in Table~\ref{tab:1}.

\begin{table}[t!]
\centering
\resizebox{\columnwidth}{!}{%
\begin{tabular}{l|ccc}
\hline
\multirow{2}{*}{} & \multicolumn{3}{c}{\textbf{The number of samples}}                                                                    \\ \cline{2-4} 
                  & \textbf{Answerable} & \textbf{\begin{tabular}[c]{@{}c@{}}Partially\\ Answerable\end{tabular}} & \textbf{Unanswerable} \\ \hline
\textbf{NetsPressoEval} & 70  & 49  & 121 \\
\textbf{TorchDataEval}  & 50  & 44  & 83  \\
\textbf{BeatNumEval}    & 78  & 59  & 140 \\
\textbf{MonkeyEval}     & 85  & 89  & 148 \\ \hline
\textbf{Total}          & 283 & 241 & 492 \\ \hline
\end{tabular}%
}
\caption{Statistics of RaCGEval}
\label{tab:1}
\end{table}

\section{Methods for Assessing Answerability}
As discussed in the introduction, the challenge of determining whether a given query in RaCG can yield a correct answer has not yet been explored. To investigate this, in this section, we describe the models we used to assess the given queries and classify their answerability.

\subsection{Zero-shot inference on instruction-following LLMs}
The first method that we use involves directly querying instruction-tuned LLMs for the answerability of each sample. To perform this zero-shot inference, we borrowed the prompt template used in LLatrieval~\cite{li-etal-2024-llatrieval}. This template was used to verify whether a given query in an RAG scenario is answerable in a general domain. The template is described in Table~\ref{tab:4} of Appendix~\ref{appendix:b}.

\subsection{Fine-tuning LLMs with automatically generated training datasets}
If we can generate training datasets for our task, we can fine-tune the parameters of the LLMs to perform answerability assessment tasks effectively. In the following sections, we describe how to construct such training datasets (\textsection 3.2.1 and \textsection 3.2.2). 

\subsubsection{Obtaining answerable samples}
In order to generate answerable samples, we need sets of API descriptions and user queries related to those APIs. To this end, we use CoNaLa \cite{yin2018conala}, which includes frequently asked queries related to the use of the Python language and high-quality solutions (i.e., valid answer code) from the StackOverflow. Although we obtained queries and solutions, there were no gold API descriptions available, so we utilized the standard Python API documentation to address this issue. We extracted the class and method keywords present in the solutions corresponding to each query and used the API descriptions associated with those keywords as the gold APIs. The process of obtaining these answerable samples is illustrated in Figure~\ref{fig:6} of Appendix~\ref{appendix:c}.

\subsubsection{Deriving partially answerable and unanswerable samples}
Using the existing answerable training samples, unanswerable and partially answerable samples can be automatically generated using the three methods discussed in Section~\ref{sec:2_3}: substituting gold APIs, concatenating topically related queries, and Substituting the query with one from out-of-database.

As a result, we were able to obtain a total of 2192 answerable, 2191 unanswerable, and 2185 partially answerable samples.

\begin{table}[t!]
\centering
\begin{tabular}{l|l|l}
\hline
\textbf{Backbone} & \textbf{Method}     & \textbf{Acc. (\%)} \\ \hline
gpt-3.5     & Zero-shot & 31.2               \\ \hline
\multirow{2}{*}{\begin{tabular}[c]{@{}c@{}}llama3\end{tabular}} & Zero-shot & 33.0 \\
                  & Fine-tuning        & 36.5               \\ \hline
\multirow{2}{*}{\begin{tabular}[c]{@{}c@{}}gemma \end{tabular}}     & Zero-shot & 36.9 \\
                  & Fine-tuning        & 46.7               \\ \hline
\end{tabular}%
\caption{Performance of answerability assessment models on RaCGEval benchmarks.}
\label{tab:2}
\end{table}

\section{Experiments}
\subsection{Experimental settings}
In this experiment, we use the following three LLMs: \textit{gemma-1.1-7b-it} \cite{team2024gemma}, \textit{llama3-instruct-8b}~\cite{dubey2024llama3herdmodels}, and \textit{gpt-3.5-turbo-0613}. For fine-tuned models, we trained them to predict one of the following three tokens for a given prompt: "A (answerable)", "U (unanswerable)" and "P (partially answerable)". Therefore, during inference, we compare the likelihoods of the three tokens to make a prediction. We employ the quantized low-Rank adaptation (QLoRA)~\cite{tim2024qlora} for parameter-efficient fine-tuning to reduce training VRAM.

\subsection{Experimental results}
We measure the performance of the answerability assessment models on the RaCGEval benchmark dataset. Table~\ref{tab:2} shows the performance of zero-shot inference or fine-tuning across various LLMs.

For zero-shot inference, the accuracy is similar to the chance level of 3-way multi-class classification across all LLM backbones. This indicates that the answerability assessment task in the code domain is not familiar to several LLMs. However, it is evident that substantial performance improvements can be achieved through fine-tuning. Nevertheless, there is still room for improvement, and we will discuss potential solutions in the discussion session.
\section{Discussion}

\begin{figure}[t]
    \centering
    \includegraphics[width=1.0\linewidth]{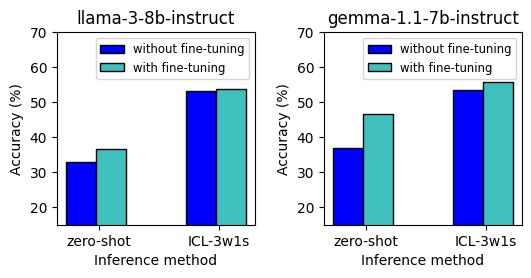}
    \caption{Effects of in-context learning for domain adaptation scenario.}
    \label{fig:4}
\end{figure}

\subsection{In-context learning for domain adaptation scenario}
We hypothesized that if we could effectively include each domain's (i.e., private libraries) information in the LLM, the performance on RaCG would improve. The in-context learning (ICL) is one simple option when only a few annotated samples are available for unseen domains. 

In addition to \textit{zero-shot} baseline, we evaluate in-context learning with 3-way-1-shot (i.e., 1-shot for each 3 classes: answerable, partially answerable, and unanswerable), denoted as ICL-3w1s, setting. The example prompt for ICL-3w1s per domain is shown in Table~\ref{tab:2shot_prompt_netspressoeval},~\ref{tab:2shot_prompt_torchdataeval},~\ref{tab:2shot_prompt_beatnumeval},~\ref{tab:2shot_prompt_monkeyeval} of Appendix D. Each set of few-shot examples is chosen randomly. 

As we can see in Figure~\ref{fig:4}, using ICL substantially improves the accuracy compared to zero-shot inference on both with and without fine-tuning. This result implies that the RaCG benchmark prioritizes LLM learning domain information (i.e., by ICL) over learning task information (i.e., by fine-tuning).

\begin{figure}[t!]
    \centering
    \includegraphics[width=1.0\linewidth]{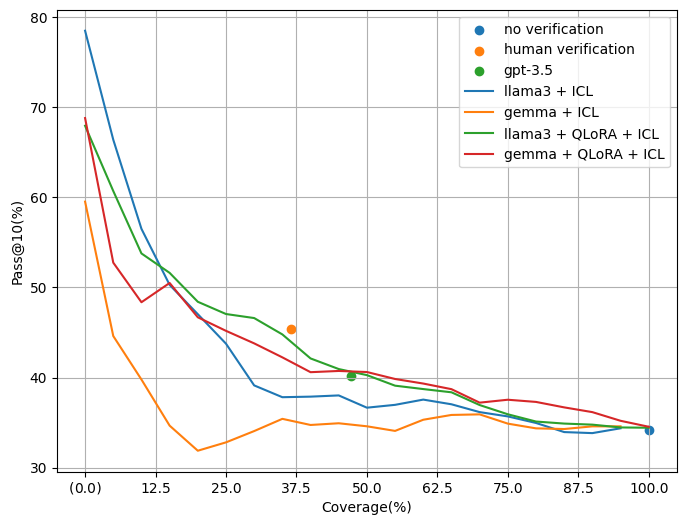}
    \caption{Trade-off between coverage and precision of generated code is explored across various answerability assessment models. We average the pass@k-values over 5\% bins of coverage.}
    \label{fig:coverage_VS_passATk}
\end{figure}

\subsection{Trade-off between coverage and precision of generated code}
We tested how much the average precision of code generation could be improved by introducing an answerability assessment. Only those samples classified as \textit{answerable} by the verification model were passed to the code generator, and \textit{unanswerable} or \textit{partially answerable} samples are not used by the generator.

We fix the code generator as the llama-3-8b-instruct model. The performance of the code generation was measured using the pass@k metric (k=10) \cite{chen2021humaneval}. Code generator sample \( n \geq k \) code given user query and code prefix, and then count the correct ones $c$ which pass the test code execution. If $n-k>=c$, pass@k is equal to $1-\prod\nolimits_{i=n-c+1}^{n} (1-\frac{k}{i})$, otherwise equals to 1. We used a version of the dataset with the canonical solution and test code added to the evaluation data by following \cite{zan-etal-2022-language}

Figure~\ref{fig:coverage_VS_passATk} shows the trade-off between coverage and pass@k. The coverage is defined as the percentage of the total test sample that is determined to be \textit{answerable}. This value can be controllable when we change the threshold of acceptance of the answerability assessment model. By setting a high/low threshold, we can force the code generation model to generate answers conservatively/aggressively.
Note that the coverage cannot be controllable in deterministic methods such as \textit{no verification}, \textit{human verification}, and API such as \textit{gpt-3.5-turbo API}\footnote{the API only provides token-wise top-k log-likelihood, which sometimes impossible to calculate log-likelihood of a whole sequence of 3-way responses}).

Among various verification models, fine-tuned (\textbf{QLoRA}) combined with in-context learning (\textbf{ICL}) on \textit{llama-3-8b-instruct} backbone show the best trade-off (i.e., the largest area under the curve (AUC)) compared to others.

\section{Conclusion}
In this work, we build and release \textbf{Retrieval-augmented Code Generability (RaCG)}, the benchmark that allows us to test the answerability of LLMs given query and retrieved APIs. \textbf{RaCG} consists of a user's query, the retrieved API descriptions related to it, the answerability assessment label (i.e., answerable/unanswerable/partially answerable), and a canonical solution. We evaluate several baselines including zero-shot, in-context learning baselines, and supervised fine-tuned models across various LLM backbones. We found that RaCG is quite challenging (i.e., best accuracy of 46.7\% with zero-shot inference) and domain adaptation is crucial to achieve higher accuracy. Moreover, we analyze the trade-off between coverage and precision of generated code when introducing the answerability assessment stage. We hope that this research will lead to the adoption of answerability assessment in various domains, and eventually improve the precision of the response of large language models.

\section*{Limitations}
Firstly, there may be more unanswerable/partially answerable types in real-world scenarios that RaCG does not cover. It would be one research direction to further define the different unanswerable types and automatically generate them from answerable samples.

Secondly, we find that domain adaptation is crucial for LLMs to achieve high accuracy on the RaCG benchmark. It would be an interesting research direction for exploring more sophisticated few-shot domain adaptation compared to in-context learning with random examples used in this work.

Third, although verification accuracy and code generation pass@k are generally proportional, the highest verification accuracy does not always achieve the best pass@k. For some queries, the LLM can generate code without the gold API documents, using its prior knowledge. However, our verification model’s annotation method does not consider this prior knowledge, which we aim to address in future research.

\section*{Acknowledgments}
This work was supported by Artificial intelligence industrial convergence cluster development project funded by the Ministry of Science and ICT(MSIT, Korea) \& Gwangju Metropolitan City.

\newpage
\bibliography{custom}

\begin{thebibliography}{17}
\providecommand{\natexlab}[1]{#1}

\bibitem[{Chen et~al.(2021)Chen, Tworek, Jun, Yuan, de~Oliveira~Pinto, Kaplan, Edwards, Burda, Joseph et~al.}]{chen2021humaneval}
Mark Chen, Jerry Tworek, Heewoo Jun, Qiming Yuan, Henrique~Ponde de~Oliveira~Pinto, Jared Kaplan, Harri Edwards, Yuri Burda, Nicholas Joseph, et~al. 2021.
\newblock Evaluating large language models trained on code.
\newblock \emph{arXiv preprint arXiv:2107.03374}.

\bibitem[{Dettmers et~al.(2023)Dettmers, Pagnoni, Holtzman, and Zettlemoyer}]{tim2024qlora}
Tim Dettmers, Artidoro Pagnoni, Ari Holtzman, and Luke Zettlemoyer. 2023.
\newblock Qlora: Efficient finetuning of quantized llms.
\newblock In \emph{Proceedings of the 37th Conference on Neural Information Processing Systems (NeurIPS 2023)}.

\bibitem[{Dubey et~al.(2024)Dubey, Jauhri, Pandey, Kadian, Al-Dahle, Letman, Mathur, Schelten, Yang, Fan et~al.}]{dubey2024llama3herdmodels}
Abhimanyu Dubey, Abhinav Jauhri, Abhinav Pandey, Abhishek Kadian, Ahmad Al-Dahle, Aiesha Letman, Akhil Mathur, Alan Schelten, Amy Yang, Angela Fan, et~al. 2024.
\newblock The llama 3 herd of models.
\newblock \emph{arXiv preprint arXiv:2407.21783}.

\bibitem[{Es et~al.(2024)Es, James, Espinosa~Anke, and Schockaert}]{esetal2024ragas}
Shahul Es, Jithin James, Luis Espinosa~Anke, and Steven Schockaert. 2024.
\newblock \href {https://aclanthology.org/2024.eacl-demo.16} {Ragas: Automated evaluation of retrieval augmented generation}.
\newblock In \emph{Proceedings of the 18th Conference of the European Chapter of the Association for Computational Linguistics: System Demonstrations}, pages 150--158, St. Julians, Malta. Association for Computational Linguistics.

\bibitem[{Fleiss(1971)}]{fleiss1971agreement}
Joseph~L. Fleiss. 1971.
\newblock Measuring nominal scale agreement among many raters.
\newblock \emph{Psychological Bulletin}, 76(5):378--382.

\bibitem[{Landis and Koch(1977)}]{landis1977kappascore}
J.~Richard Landis and Gary~G. Koch. 1977.
\newblock The measurement of observer agreement for categorical data.
\newblock \emph{Biometrics}.

\bibitem[{Li et~al.(2023)Li, Cheng, Zhao, Nie, and Wen}]{li2023halueval}
Junyi Li, Xiaoxue Cheng, Xin Zhao, Jian-Yun Nie, and Ji-Rong Wen. 2023.
\newblock \href {https://doi.org/10.18653/v1/2023.emnlp-main.397} {{H}alu{E}val: A large-scale hallucination evaluation benchmark for large language models}.
\newblock In \emph{Proceedings of the 2023 Conference on Empirical Methods in Natural Language Processing}, pages 6449--6464, Singapore. Association for Computational Linguistics.

\bibitem[{Li et~al.(2024)Li, Zhu, Li, Yin, Sun, and Qiu}]{li-etal-2024-llatrieval}
Xiaonan Li, Changtai Zhu, Linyang Li, Zhangyue Yin, Tianxiang Sun, and Xipeng Qiu. 2024.
\newblock \href {https://doi.org/10.18653/v1/2024.naacl-long.305} {{LL}atrieval: {LLM}-verified retrieval for verifiable generation}.
\newblock In \emph{Proceedings of the 2024 Conference of the North American Chapter of the Association for Computational Linguistics: Human Language Technologies (Volume 1: Long Papers)}, pages 5453--5471, Mexico City, Mexico. Association for Computational Linguistics.

\bibitem[{Manakul et~al.(2023)Manakul, Liusie, and Gales}]{manakul2023selfcheckgpt}
Potsawee Manakul, Adian Liusie, and Mark Gales. 2023.
\newblock \href {https://doi.org/10.18653/v1/2023.emnlp-main.557} {{S}elf{C}heck{GPT}: Zero-resource black-box hallucination detection for generative large language models}.
\newblock In \emph{Proceedings of the 2023 Conference on Empirical Methods in Natural Language Processing}, pages 9004--9017, Singapore. Association for Computational Linguistics.

\bibitem[{Mesnard et~al.(2024)Mesnard, Hardin, Dadashi, Bhupatiraju, Pathak, Sifre, Rivi{\`e}re, Kale, Love et~al.}]{team2024gemma}
Thomas Mesnard, Cassidy Hardin, Robert Dadashi, Surya Bhupatiraju, Shreya Pathak, Laurent Sifre, Morgane Rivi{\`e}re, Mihir~Sanjay Kale, Juliette Love, et~al. 2024.
\newblock Gemma: Open models based on gemini research and technology.
\newblock \emph{arXiv preprint arXiv:2403.08295}.

\bibitem[{{Nota Inc.}(2024)}]{netspresso2024}
{Nota Inc.} 2024.
\newblock Netspresso api documentation.
\newblock \url{https://nota-netspresso.github.io/PyNetsPresso/}.

\bibitem[{Shuster et~al.(2021)Shuster, Poff, Chen, Kiela, and Weston}]{shuster2021RAGreducehalu}
Kurt Shuster, Spencer Poff, Moya Chen, Douwe Kiela, and Jason Weston. 2021.
\newblock \href {https://doi.org/10.18653/v1/2021.findings-emnlp.320} {Retrieval augmentation reduces hallucination in conversation}.
\newblock In \emph{Findings of the Association for Computational Linguistics: EMNLP 2021}, pages 3784--3803, Punta Cana, Dominican Republic. Association for Computational Linguistics.

\bibitem[{Tian et~al.(2024)Tian, Yan, Yang, Chen, Wang, Luo, and Ma}]{tian2024codehalu}
Yuchen Tian, Weixiang Yan, Qian Yang, Qian Chen, Wen Wang, Ziyang Luo, and Lei Ma. 2024.
\newblock \href {https://arxiv.org/abs/2405.00253} {Codehalu: Code hallucinations in llms driven by execution-based verification}.
\newblock \emph{Preprint}, arXiv:2405.00253.

\bibitem[{Wang et~al.(2023)Wang, Le, Gotmare, Bui, Li, and Hoi}]{wang-etal-2023-codet5}
Yue Wang, Hung Le, Akhilesh Gotmare, Nghi Bui, Junnan Li, and Steven Hoi. 2023.
\newblock \href {https://doi.org/10.18653/v1/2023.emnlp-main.68} {{C}ode{T}5+: Open code large language models for code understanding and generation}.
\newblock In \emph{Proceedings of the 2023 Conference on Empirical Methods in Natural Language Processing}, pages 1069--1088, Singapore. Association for Computational Linguistics.

\bibitem[{Yin et~al.(2018)Yin, Deng, Chen, Vasilescu, and Neubig}]{yin2018conala}
Pengcheng Yin, Bowen Deng, Edgar Chen, Bogdan Vasilescu, and Graham Neubig. 2018.
\newblock Learning to mine aligned code and natural language pairs from stack overflow.
\newblock In \emph{International Conference on Mining Software Repositories}, MSR, pages 476--486. ACM.

\bibitem[{Zan et~al.(2023)Zan, Bei~Chen, Cao, Zhang, Wu, Guan, Yin, , and Wang}]{zan2023privatelibrary}
Daoguang Zan, Yongshun~Gong Bei~Chen, Junzhi Cao, Fengji Zhang, Bingchao Wu, Bei Guan, Yilong Yin, , and Yongji Wang. 2023.
\newblock Private-library-oriented code generation with large language models.
\newblock \emph{arXiv preprint arxiv:2307.15370}.

\bibitem[{Zan et~al.(2022)Zan, Chen, Lin, Guan, Yongji, and Lou}]{zan-etal-2022-language}
Daoguang Zan, Bei Chen, Zeqi Lin, Bei Guan, Wang Yongji, and Jian-Guang Lou. 2022.
\newblock \href {https://doi.org/10.18653/v1/2022.findings-emnlp.21} {When language model meets private library}.
\newblock In \emph{Findings of the Association for Computational Linguistics: EMNLP 2022}, pages 277--288, Abu Dhabi, United Arab Emirates. Association for Computational Linguistics.

\end{thebibliography}

\newpage
\appendix

\onecolumn
\section{An Example of NetsPressoEval Dataset}
\label{appendix:a}
\begin{table*}[!h]
\centering
\begin{tabular}{|cl|l|}
\hline
\multicolumn{2}{|c|}{\begin{tabular}[c]{@{}l@{}}query\end{tabular}} &
  \begin{tabular}[c]{@{}l@{}}Can you guide me through compressing my custom model, retraining it, \\and measuring its latency?\end{tabular} \\ \hline
\multicolumn{1}{|c|}{\multirow{3}{*}{\begin{tabular}[c]{@{}c@{}}gold\\ APIs\end{tabular}}} &
  1 &
  Compressor.automatic\_compression(PARAMETERS): DESCRIPTION \\ \cline{2-3} 
\multicolumn{1}{|c|}{} &
  2 &
  Trainer.train(PARAMETERS): DESCRIPTION \\ \cline{2-3} 
\multicolumn{1}{|c|}{} &
  3 &
  Benchmarker.benchmark\_model(PARAMETERS): DESCRIPTION \\ \hline
\multicolumn{2}{|l|}{\begin{tabular}[c]{@{}l@{}}canonical\\ solution\end{tabular}} &
  \begin{tabular}[c]{@{}l@{}}from netspresso import NetsPresso \\ from netspresso.trainer.optimizers import AdamW \\ from netspresso.enums import DeviceName, SoftwareVersion\\ \\ netspresso = NetsPresso(email='YOUR\_EMAIL', password='YOUR\_PASSWORD')  \\ \\ \# Compress\\ compressor = netspresso.compressor()\\ compressed\_model = compressor.automatic\_compression(\\    \quad\quad input\_shapes={[}\{"batch": 1, "channel": 3, "dimension": {[}224, 224{]}\}{]},\\    \quad\quad input\_model\_path="./models/YOUR\_PYTORCH\_MODEL.pt",\\     \quad\quad output\_dir="./outputs/compressed", compression\_ratio=0.5,\\ )\\ \\ \# Retrain\\ trainer = netspresso.trainer(yaml\_path="hparams.yaml")\\ trainer.set\_fx\_model(fx\_model\_path="YOUR\_FX\_MODEL.pt") \\ trainer.set\_training\_config(epochs=100, batch\_size=128, optimizer=AdamW(lr=1e-4))\\ trainer.train(gpus="0, 1", project\_name="project\_retrain")\\ \\ \# Benchmark\\  benchmarker = netspresso.benchmarker()\\  benchmark\_result = benchmarker.benchmark\_model(\\     \quad\quad input\_model\_path="YOUR\_RETRAINED\_MODEL\_PATH", \\     \quad\quad target\_device\_name=DeviceName.JETSON\_AGX\_ORIN, \\     \quad\quad target\_software\_version=SoftwareVersion.JETPACK\_5\_0\_1,\\ )\end{tabular} \\ \hline
\end{tabular}
\caption{Example of NetsPressoEval dataset.}
\label{tab:3}
\end{table*}

\newpage
\section{Zero-shot Inference and In-Context Learning Prompt Template}
\label{appendix:b}
\begin{table*}[h!]
\centering

\begin{tabular}{|l|}
\hline
\begin{tabular}[c]{@{}l@{}}\# Profile\\ - Role: Answerability Assessment GPT, evaluating whether provided documents can support answering\\  the user's questions effectively.\\ \\ \# Demonstration\\ \#\# Demo 1\\ Question: \{QUESTION\}\\ Candidate Documents\\ \{\{APIs\}\}\\ Judgement: \{JUDGEMENT\}\\ ...\\ \# Input\\ - Question: User's specific question(s).\\ - Candidate Documents: Documents potentially useful for answering the questions.\\ \\ \# Skill\\ 1. Analyzing questions to understand required information.\\ 2. Assessing documents to determine their ability to support clear and accurate answers.\\ \\ \# Rules\\ 1. Maintain character.\\ 2. Provide final verdict only as "{[}Answerable{]}", "{[}Partially answerable{]}", or "{[}Unanswerable{]}".\\ 3. Do not evaluate demonstration, only assess question(s) and documents.\\ 4. Adhere strictly to output format.\\ \\ \# Judgment Criteria\\ 1. Document length should not influence evaluation.\\ 2. Strive for objectivity.\\ 3. "{[}Answerable{]}" if documents support clear, accurate, and engaging answers, "{[}Partially answerable{]}" \\ if some aspects can be answered, "{[}Unanswerable{]}" if no relevant information.\\ \\ \# Workflow\\ 1. Understand user questions.\\ 2. Evaluate documents for answer support.\\ 3. Provide final judgment.\\ \\ \# Reminder\\ Always remember the role settings.\\ \\ Question: \{\{query\}\}\\ \\ Candidate Documents\\ \{\{apis\}\}\\ \\ Judgment: '''\end{tabular} \\ \hline
\end{tabular}
\caption{In-context learning prompt format. When performing zero-shot inference, the demonstration section is not included in the input prompt. The prompt templates are mostly adapted from LLatrieval~\cite{li-etal-2024-llatrieval}.}
\label{tab:4}

\end{table*}

\newpage
\section{Training Gold API Dataset Construction}
\label{appendix:c}

The following figure is the method to construct gold APIs for the training dataset.

\begin{figure*}[ht!]
    \centering
    \includegraphics[width=1.0\linewidth]{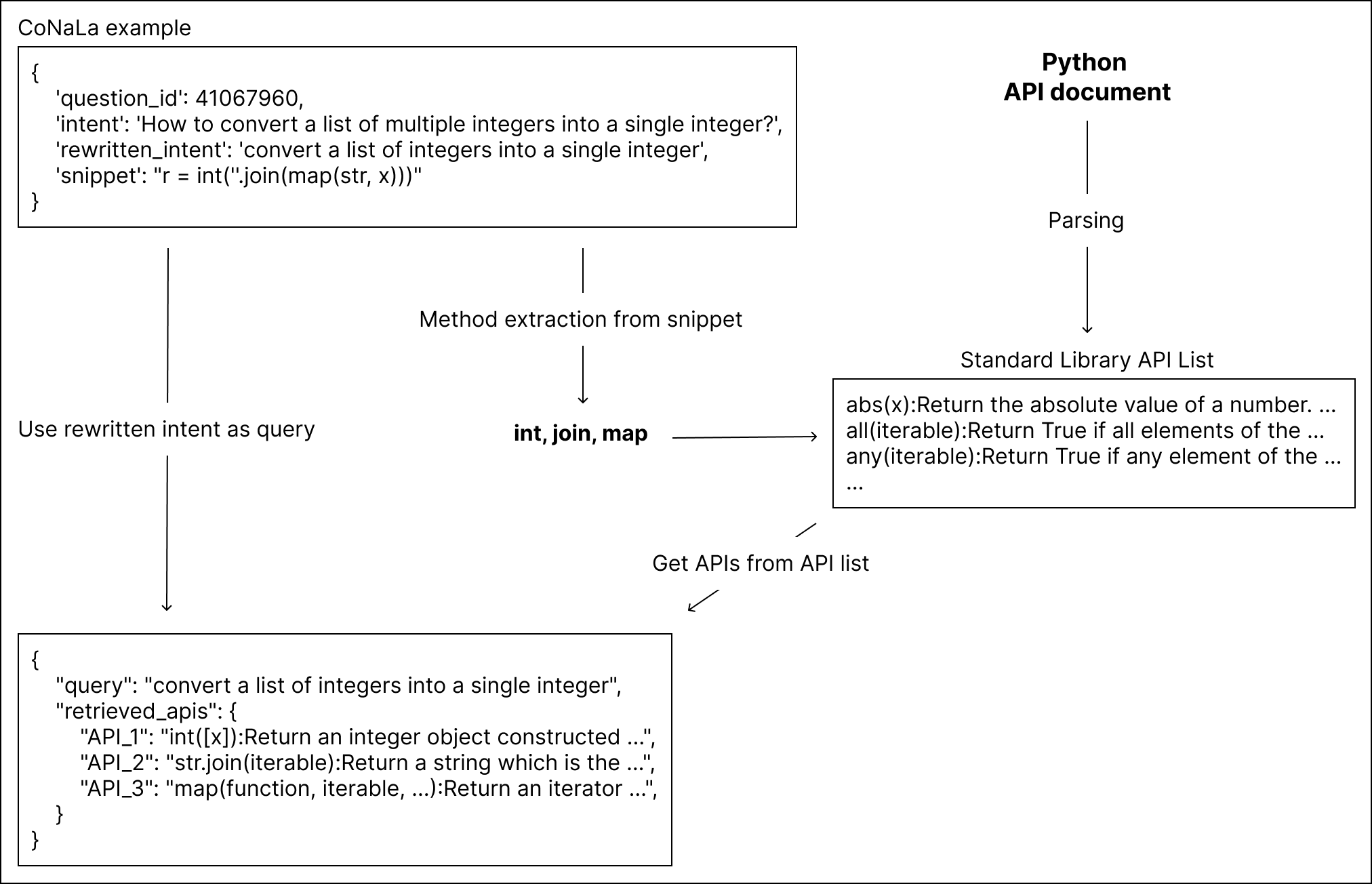}
    \caption{Standard Python APIs matched with CoNaLa code generation query.}
    \label{fig:6}
\end{figure*}

\newpage

\section{Few-shot samples chosen per domain}
The following tables show examples of 3-way-1-shot samples for in-context learning. 

\begin{table*}[h!]
\small  
\centering
\begin{tabularx}{\textwidth}{|X|}
\hline
\#\#\# Demo 1\\ Question: How can I compress a custom ONNX model using singular value decomposition?\\ \\ Candidate Documents\\ API\_1: Converter.convert\_model(input\_model\_path: str, output\_dir: str, target\_framework: Union{[}str, Framework{]}, target\_device\_name: Union{[}str, DeviceName{]}, target\_data\_type: Union{[}str, DataType{]} = DataType.FP16, target\_software\_version: Optional{[}Union{[}str, SoftwareVersion{]}{]} = None, input\_shape: Optional{[}InputShape{]} = None, dataset\_path: Optional{[}str{]} = None, wait\_until\_done: bool = True): Convert a model to the specified framework. Returns model conversion task dictionary.\\ API\_2: Compressor.recommendation\_compression(compression\_method: CompressionMethod, recommendation\_method: RecommendationMethod, recommendation\_ratio: float, input\_model\_path: str, output\_dir: str, input\_shapes: List{[}Dict{[}str, int{]}{]}, framework: Framework = Framework.PYTORCH, options: Options = Options(), dataset\_path: Optional{[}str{]} = None): Compress a recommendation-based model using the given compression and recommendation methods. Returns source model and compressed model information.\\ API\_3: Compressor.automatic\_compression(input\_model\_path: str, output\_dir: str, input\_shapes: List{[}Dict{[}str, int{]}{]}, framework: Framework = Framework.PYTORCH, compression\_ratio: float = 0.5): Compress a model automatically based on the given compression ratio.\\ \\ Judgment: A\\ \\ By using above documents, you can compress the model using API\_2 Compressor.recommendation\_compression().\\ As documents support clear, accurate, and engaging answers, query is answerable so your output should be A.\\ \\ \#\#\# Demo 2\\ Question: Show me an example of training a ResNet model for image classification.\\ \\ Candidate Documents\\ API\_1: Trainer.set\_model\_config(model\_name: str, img\_size: int, use\_pretrained: bool = True, load\_head: bool = False, path: Optional{[}str{]} = None, fx\_model\_path: Optional{[}str{]} = None, optimizer\_path: Optional{[}str{]} = None): Set the model configuration for the Trainer.\\ API\_2: Trainer.set\_logging\_config(project\_id: Optional{[}str{]} = None, output\_dir: str = './outputs', tensorboard: bool = True, csv: bool = False, image: bool = True, stdout: bool = True, save\_optimizer\_state: bool = True, validation\_epoch: int = 10, save\_checkpoint\_epoch: Optional{[}int{]} = None): Set the logging configuration.\\ API\_3: Trainer.set\_training\_config(optimizer, scheduler, epochs: int = 3, batch\_size: int = 8): Set the training configuration.\\ \\ Judgment: P\\ \\ You can set ResNet model using API\_1 Trainer.set\_model\_config(), but you cannot select the task with above documents.\\ You can partially answer to the query, so your output should be P.\\ \\ \#\#\# Demo 3\\ Question: How do I set random seed for environmental setup?\\ \\ Candidate Documents\\ API\_1: Trainer.set\_logging\_config(project\_id: Optional{[}str{]} = None, output\_dir: str = './outputs', tensorboard: bool = True, csv: bool = False, image: bool = True, stdout: bool = True, save\_optimizer\_state: bool = True, validation\_epoch: int = 10, save\_checkpoint\_epoch: Optional{[}int{]} = None): Set the logging configuration.\\ API\_2: Trainer.train(gpus: str, project\_name: str): Train the model with the specified configuration. Returns a dictionary containing information about the training.\textbackslash{}n\textbackslash{}n\textbackslash{}nParameters\textbackslash{}n----------\textbackslash{}ngpus (str): GPU ids to use, separated by commas.\textbackslash{}nproject\_name (str): Project name to save the experiment.\\ API\_3: Trainer.set\_training\_config(optimizer, scheduler, epochs: int = 3, batch\_size: int = 8): Set the training configuration.\\ \\ Judgment: U\\ \\ You may need information of methods like 'set\_environment\_config()' to resolve the question.\\ But there's no relevant information about the methods. You are unanswerble to the query, so your output should be U. \\ \hline
\end{tabularx}
\caption{3-way-1shot example for NetsPressoEval}
\label{tab:2shot_prompt_netspressoeval}
\end{table*}

\newpage
\begin{table*}[t!]
\small  
\centering
\begin{tabularx}{\textwidth}{|X|}
\hline
\#\#\# Demo 1\\ Question: How to augument the datapipe by repeating it six times.\\ \\ Candidate Documents\\ API\_1: map(*args, **kwds):\textbackslash{}n    Apply the input function over each item from the source DataPipe (functional name: ``map``).\textbackslash{}n    The function can be any regular Python function or partial object.\\ API\_2: EndOnDiskCacheHolder(datapipe, mode='wb', filepath\_fn=None, *, same\_filepath\_fn=False, skip\_read=False):\textbackslash{}n    Indicates when the result of prior DataPipe will be saved local files specified\textbackslash{}n    by ``filepath\_fn`` (functional name: ``end\_caching``). Moreover, the result of source DataPipe\textbackslash{}n    is required to be a tuple of metadata and data, or a tuple of metadata and file handle.\\ API\_3: cycle(*args, **kwds):\textbackslash{}n    Cycles the specified input in perpetuity by default, or for the specified number of times (functional name: ``cycle``).\\ \\ Judgment: A\\ \\ By using above documents, you can generate the code snippet like 'datapipe.cycle(6)'.\\ As documents support clear, accurate, and engaging answers, query is answerable so your output should be A.\\ \\ \#\#\# Demo 2\\ Question: Read the URL using the HTTP protocol and process the csv file.\\ \\ Candidate Documents\\ API\_1: IoPathFileLister(*args, **kwds):\textbackslash{}n    Lists the contents of the directory at the provided ``root`` pathname or URL, and yields the full pathname or URL for each file within the directory.\\ API\_2: OnlineReader(*args, **kwds):\textbackslash{}n    Takes file URLs (can be HTTP URLs pointing to files or URLs to GDrive files), and yields tuples of file URL and IO stream.\\ API\_3: HttpReader(source\_datapipe: IterDataPipe{[}str{]}, timeout: Optional{[}float{]} = None):\textbackslash{}n    Takes file URLs (HTTP URLs pointing to files), and yields tuples of file URL and IO stream.\\ \\ Judgment: P\\ \\ You can read the URL with API\_3 HttpReader, but you cannot process the csv file with above documents.\\ You can partially answer to the query, so your output should be P.\\ \\ \#\#\# Demo 3\\ Question: Clone the source datapipe two times\\ \\ Candidate Documents\\ API\_1: unbatch(datapipe: IterDataPipe, unbatch\_level: int = 1):\textbackslash{}n    Undoes batching of data (functional name: ``unbatch``). In other words, it flattens the data up to the specified level within a batched DataPipe.\\ API\_2: IterToMapConverter(*args, **kwds):\textbackslash{}n    Lazily load data from ``IterDataPipe`` to construct a ``MapDataPipe`` with the key-value pair generated by ``key\_value\_fn`` (functional name: ``to\_map\_datapipe``).\\ API\_3: cycle(*args, **kwds):\textbackslash{}n    Cycles the specified input in perpetuity by default, or for the specified number of times (functional name: ``cycle``).\\ \\ Judgment: U\\ \\ You may need information of methods like 'clone()' to resolve the question.\\ But there's no relevant information about the methods. You are unanswerble to the query, so your output should be U. \\ \hline
\end{tabularx}
\caption{3-way-1shot example for TorchDataEval}
\label{tab:2shot_prompt_torchdataeval}
\end{table*}

\newpage
\begin{table*}[t!]
\small  
\centering
\begin{tabularx}{\textwidth}{|X|}
\hline
\#\#\# Demo 1\\ Question: create a beatnum numset composed of a list {[}{[}8, 7, 2{]}, {[}5, 6, 1{]}, {[}8, 2, 6{]}{]}\\ \\ Candidate Documents\\ API\_1: numset(obj, itemsize=None, copy=True, unicode=None, order=None):\textbackslash{}n    Create a `charnumset`.\\ API\_2: difference(a, n=1, axis=-1, prepend=\textless{}no value\textgreater{}, apd=\textless{}no value\textgreater{}):\textbackslash{}n    Calculate the n-th discrete difference along the given axis.\\ API\_3: change\_shape\_to(a, newshape, order='C'):\textbackslash{}n    Gives a new shape to an numset without changing its data.\\ \\ Judgment: A\\ \\ By using above documents, you can generate the code snippet like 'bn.numset({[}{[}8, 7, 2{]}, {[}5, 6, 1{]}, {[}8, 2, 6{]}{]})'.\\ As documents support clear, accurate, and engaging answers, query is answerable so your output should be A.\\ \\ \#\#\# Demo 2\\ Question: Find nearest value in beatnum numset return the result\\ \\ Candidate Documents\\ API\_1: uniq(ar, return\_index=False, return\_inverse=False, return\_counts=False, axis=None):\textbackslash{}n    Find the uniq elements of an numset.\\ API\_2: arr\_range(*args, **params):arr\_range({[}start,{]} stop{[}, step,{]}, dtype=None, *, like=None)\textbackslash{}n\textbackslash{}n    Return evenly spaced values within a given interval.\\ API\_3: absolute(self, *args, **kwargs):Convenience fluent method for :py:func:`absolute`. The arguments are the same as for :py:func:`absolute`, with this numset as data.\\ \\ Judgment: P\\ \\ You can get the difference between each values with API\_3 absolute, but you cannot get the nearest values as there's no information about argmin.\\ You can partially answer to the query, so your output should be P.\\ \\ \#\#\# Demo 3\\ Question: Convert beatnum numset type and values from Float64 to Float32\\ \\ Candidate Documents\\ API\_1: difference(a, n=1, axis=-1, prepend=\textless{}no value\textgreater{}, apd=\textless{}no value\textgreater{}):\textbackslash{}n    Calculate the n-th discrete differenceerence along the given axis.\\ API\_2: average(a, axis=None, dtype=None, out=None, keepdims=False):\textbackslash{}n    Compute the arithmetic average along the specified axis. Returns the average of the numset elements.\\ API\_3: total\_count(a, axis=None, dtype=None, out=None, keepdims=\textless{}no value\textgreater{}, initial=\textless{}no value\textgreater{}, where=\textless{}no value\textgreater{}):\textbackslash{}n    Sum of numset elements over a given axis.\\ \\ Judgment: U\\ \\ You may need information of methods like 'convert\_type()' to resolve the question.\\ But there's no relevant information about the methods. You are unanswerble to the query, so your output should be U. \\ \hline
\end{tabularx}
\caption{3-way-1shot example for BeatNumEval}
\label{tab:2shot_prompt_beatnumeval}
\end{table*}

\newpage
\begin{table*}[t!]
\small  
\centering
\begin{tabularx}{\textwidth}{|X|}
\hline
\#\#\# Demo 1\\ Question: deleting a column from a Monkey KnowledgeFrame return the changged knowledgeframe\\ \\ Candidate Documents\\ API\_1: choose\_dtypes(self, include=None, exclude=None) -\textgreater 'KnowledgeFrame':\textbackslash{}n        Return a subset of the KnowledgeFrame's columns based on the column dtypes.\\ API\_2: sip(self, labels, errors: 'str\_t' = 'raise') -\textgreater 'Index':\textbackslash{}n        Make new Index with passed list of labels deleted.\\ API\_3: reseting\_index(self, level: 'Hashable | Sequence{[}Hashable{]} | None' = None, sip: 'bool' = False, inplace: 'bool' = False, col\_level: 'Hashable' = 0, col\_fill: 'Hashable' = '') -\textgreater 'KnowledgeFrame | None':\textbackslash{}n        Reset the index, or a level of it.\\ \\ Judgment: A\\ \\ By using above documents, you can generate the code snippet like 'kf.sip(column\_name, axis=1)'.\\ As documents support clear, accurate, and engaging answers, query is answerable so your output should be A.\\ \\ \#\#\# Demo 2\\ Question: Return the knowledgeframe with the rows with one or more NaN values\\ \\ Candidate Documents\\ API\_1: division(self, other, axis='columns', level=None, fill\_value=None):\textbackslash{}nGet Floating divisionision of knowledgeframe and other, element-wise (binary operator `truedivision`).\\ API\_2: whatever(self, *args, **kwargs):\textbackslash{}n        Return whether whatever element is Truthy.\\ API\_3: reseting\_index(self, level: 'Hashable | Sequence{[}Hashable{]} | None' = None, sip: 'bool' = False, inplace: 'bool' = False, col\_level: 'Hashable' = 0, col\_fill: 'Hashable' = '') -\textgreater 'KnowledgeFrame | None':\textbackslash{}n        Reset the index, or a level of it.\\ \\ Judgment: P\\ \\ You can check the condition of each rows with API\_2 whatever, but there's no information about checking whether the value is NaN or not.\\ You can partially answer to the query, so your output should be P.\\ \\ \#\#\# Demo 3\\ Question: I want to make all column headers in my monkey data frame lower case\\ \\ Candidate Documents\\ API\_1: KnowledgeFrame(data=None, index: 'Axes | None' = None, columns: 'Axes | None' = None, dtype: 'Dtype | None' = None, clone: 'bool | None' = None):\textbackslash{}n    Two-dimensional, size-mutable, potentitotal\_ally heterogeneous tabular data.\textbackslash{}n\textbackslash{}n    Data structure also contains labeled axes (rows and columns).\\ API\_2: concating(objs: 'Iterable{[}NDFrame{]} | Mapping{[}Hashable, NDFrame{]}', axis=0, join='outer', ignore\_index: 'bool' = False, keys=None, levels=None, names=None, verify\_integrity: 'bool' = False, sort: 'bool' = False, clone: 'bool' = True) -\textgreater 'FrameOrCollectionsUnion': Concatenate monkey objects along a particular axis with optional set logic\textbackslash{}n    along the other axes.\\ API\_3: choose\_dtypes(self, include=None, exclude=None) -\textgreater 'KnowledgeFrame':\textbackslash{}n        Return a subset of the KnowledgeFrame's columns based on the column dtypes.\\ \\ Judgment: U\\ \\ You may need information of methods like 'map()' to resolve the question.\\ But there's no relevant information about the methods. You are unanswerble to the query, so your output should be U. \\ \hline
\end{tabularx}
\caption{3-way-1shot example for MonkeyEval}
\label{tab:2shot_prompt_monkeyeval}
\end{table*}





\end{document}